# Gliding in extreme waters: Dynamic Modeling and Nonlinear Control of an Agile Underwater Glider

**Hanzhi Yang**\*, **Nina Mahmoudian**\*

\* *Department of Mechanical Engineering, Purdue University, West Lafayette, IN 47907 USA (E-mail: yang1118, ninam@purdue.edu)*

**Abstract:** This paper describes the modeling of a custom-made underwater glider capable of flexible maneuvers in constrained areas and proposes a control system. Due to the lack of external actuators, underwater gliders can be greatly influenced by environmental disturbance. In addition, the nonlinearity of the system affects the motions during the transition between each flight segment. Here, a data-driven parameter estimation experimental methodology is proposed to identify the nonlinear dynamics model for our underwater glider using an underwater motion capture system. Then, a nonlinear system controller is designed based on Lyapunov function to overcome environmental disturbance, potential modeling errors, and nonlinearity during flight state transitions. The capability of lowering the impact of environmental disturbance is validated in simulations. A hybrid control system applying PID controller to maintain steady state flights and the proposed controller to switch between states is also demonstrated by performing complex maneuvers in simulation. The proposed control system can be applied to gliders for reliable navigation in dynamic water areas such as fjords where the sea conditions may vary from calm to rough seasonally.

## 1. INTRODUCTION

Autonomous underwater vehicles (AUV) are widely used to conduct ocean explorations. As a type of AUV with long endurance missions, underwater glider (UG) has become one of the most effective underwater unmanned vehicles for submerged surveys in recent years. Unlike other AUVs, UG achieves gliding motions by altering its net buoyancy and center of gravity, which gives it advantages of long endurance, low energy consumption, low noise, etc. As more and more UG models have been developed to operate in dynamic and complex underwater environments, motion control is one of the most important features for ensuring reliable and robust underwater operations of a glider. Many classic glider models applied proportional-integral-derivative (PID) controller for heading or pitch control (Eriksen et al. (2001); Sherman et al. (2001); Webb et al. (2001)), yet the cooperations of multiple internal actuation modules have limited the control accuracy and efficiency and therefore simple linear controllers have limitations dealing with complex missions especially when more advanced path planning algorithms are implemented. In addition, the uncertainty and modeling errors in the glider dynamics system need to be taken into consideration for controller development to ensure the reliability and robustness of the underwater vehicle system.

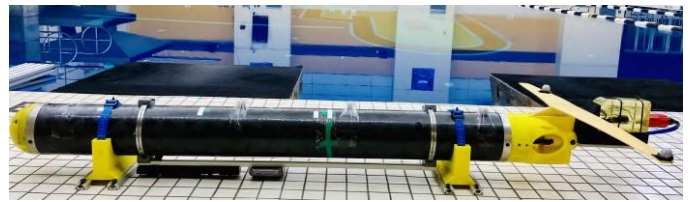

Fig. 1. Underwater glider ROUGHIE with tracking markers attached as described in section 2; the trim weights are visible at the bottom

The Research Oriented Underwater Glider for Hands-on Investigative Engineering (ROUGHIE) (as shown in Fig. 1) is a low-cost and modular prototype for extending the maneuverability of UGs in constrained spaces like near shore shallow water areas. It is 1.2 *m* long and weighs 13 *kg*. The small size and weight enable the glider shore launch, operation in shallow environments, and indoor pool operation. In our previous work (Lambert et al. (2022)), a feedforward-PID controller is designed for both pitch and roll controls, and an on-off controller is used for depth control. The closed-loop control methodology was applied during the steady-state flights while during the transition between flight segments, the glider used an open-loop control. This linear control system has been tested in a diving pool to show the glider's maneuverability in a constrained environment. To further deploy the glider in an outdoor water area where the waves and currents introduce disturbance to the system, affecting the glider's flights, an advanced control system for transiting between flight segments is needed. Such advanced controller requires the knowledge of the mathematical model of the vehicle.

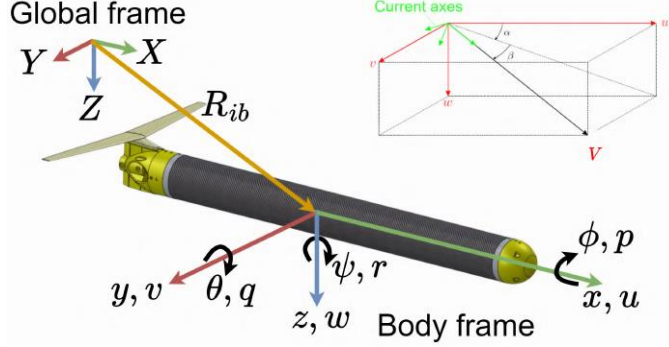

Fig. 2. 3D illusion of ROUGHIE and its coordinate system for dynamics analysis

In this paper, we presented two works: first, we showed our experimental method, utilizing Qualisys underwater motion capture cameras, of UG dynamics modeling for our custom-made glider, and second, we designed and implemented a nonlinear system controller for the glider. The data-driven modeling approach applied the Markov Chain Monte Carlo (MCMC) method to estimate multiple hydrodynamic parameters. The MCMC method, which has been applied for parameter estimation based on imperfect data (Kusari et al. (2022)), uses sequential sampling starting from a prior probability distribution and gradually moves toward the target distribution. We specifically utilized the Metropolis-Hastings (M-H) algorithm to determine the distributions of unknown parameters in the UG dynamics equation of motions using experimental data, and then used these distributions to design a nonlinear controller (NLC) for the glider on depth and pitch controls to overcome disturbance and modeling errors. The methods and algorithms in this work are validated in simulations. In the end, a hybrid control system involving PID for maintaining trim states and NLC for transiting between flight segments is applied on the glider to perform circular and S-curved maneuvering patterns.

This paper is set up as follows: Section 2 reviews the dynamics model of the glider system and presents the data driven parameter estimation experimental methodology; Section 3 proposes a NLC using the estimated system model; Section 4 validates the controller in simulations; Section 5 concludes the work and considers future works.

## 2. DYNAMICS MODEL AND PARAMETER ESTIMATION

This section provides a brief analysis of the glider dynamics system and the method of parameter estimation for system identification.

### 2.1 Dynamics system of underwater gliders

In a previous study (Ziaeefard et al. (2018)), an analysis of kinematics and dynamics of the custom-made underwater glider ROUGHIE was conducted (in a coordinate system shown in Fig. 2), and the dynamics model is given by

$$\dot{\nu} = M^{-1}(-\dot{M}\nu + \begin{bmatrix} P \times \Omega \\ Q \times \Omega + P \times V \end{bmatrix} \\ + \begin{bmatrix} m_b g(R_{ib})^T \hat{k} \\ (m_r r_r + m_s r_s + m_b r_b)g \times (R_{ib})^T \hat{k} \end{bmatrix} \\ + \begin{bmatrix} F_{ext} \\ T_{ext} \end{bmatrix} \quad (1)$$

in which $\nu = [V^T, \Omega^T]^T = [u,v,w,p,q,r]^T$ is a velocity vector containing translational ($V$) and rotational ($\Omega$) velocities in body frame, M is the inertia matrix, and P and Q are the translational and angular momentum in body frame. Rotational matrix $R_{ib}$ maps global frame to body frame following Euler angle sequence $\{\psi\theta\phi\}$ and is given by

$$R_{ib} = \begin{bmatrix} c\theta c\psi & s\phi s\theta c\psi - c\phi s\psi & c\phi s\theta c\psi + s\phi s\psi \\ c\theta s\psi & c\phi c\psi + s\phi s\theta s\psi & -s\phi c\psi + c\phi s\theta s\psi \\ -s\theta & s\phi c\theta & c\phi c\theta \end{bmatrix} \quad (2)$$

using the notation $s\cdot$ as $sin(\cdot)$ and $c\cdot$ as $cos(\cdot)$.

$F_{ext}$ and $T_{ext}$ are the external forces and torques acting on the glider in the body frame, and the rotational matrix mapping flow frame to body frame $R_{bf}$ is defined using the angle of attack $\alpha = atan2(w,u)$ and sideslip angle $\beta = sin^{-1}(\frac{v}{\sqrt{u^2+v^2+w^2}})$ and is expressed as

$$R_{bf} = \begin{bmatrix} c\alpha c\beta & -c\alpha s\beta & -s\alpha \\ s\beta & c\beta & 0 \\ s\alpha c\beta & -s\alpha s\beta & c\alpha \end{bmatrix}. \quad (3)$$

The external forces and torques can be calculated using twelve hydrodynamics coefficients K's

$$F_{ext} = [F_1, F_2, F_3]^T = R_{bf}[-D, SF, -L]^T$$

$$T_{ext} = [T_1, T_2, T_3]^T = R_{bf}[T_{DL1}, T_{DL2}, T_{DL3}]^T$$

$$D = (K_{D0} + K_D \alpha^2)V^2$$

$$L = (K_{L0} + K_L \alpha^2)V^2 \quad (4)$$

$$SF = K_\beta \beta V^2$$

$$T_{DL1} = (K_{MR}\beta + K_p p)V^2$$

$$T_{DL2} = (K_{M0} + K_M \alpha + K_q q)V^2 \quad T_{DL3} = (K_{MY}\beta + K_r r)V^2.$$

## 2.2 Experiment setup for system identification

The parameters to be identified in this system can be categorized into two groups: first, measurable parameters include mass $m_t$ and moments of inertia $I_{3\times 3}$, which form the inertia matrix M (added mass is estimated based on the vehicle's geometry), masses of moving components $[m_r,m_s,m_b]^T$, and their positions $[r_r,r_s,r_b]^T$ inside the hull; second, parameters that need to be estimated mathematically include all the $K$'s which are the hydrodynamics coefficients.

In some other works (such as Graver et al. (2003); Zhang et al. (2013)), the values of the hydrodynamics coefficients $K$'s are estimated using fluid physics simulation software due to the vehicles' size and operation depth and therefore the incapability of conducting indoor experiments. ROUGHIE's small size and light weight allows it to operate in small indoor water areas like a swimming

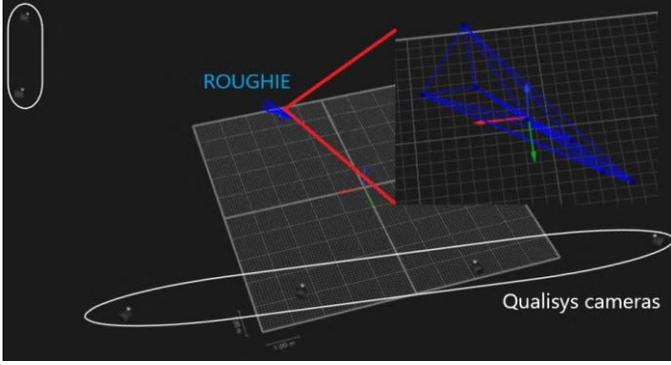

Fig. 3. Qualisys motion capture UI. Six underwater motion capture cameras were used to track the glider, shown as a blue wire-frame, each node representing one marker

pool. Hence, in this project, the coefficients are estimated through experiments conducted at a diving pool in Morgan J. Burke Aquatic Center at Purdue University. A $Qualisys^{TM}$ underwater motion capture system was used to measure and record the real-time glider positions and orientations under various combinations of control inputs by tracking the markers stuck on the surface of the glider (as shown in Fig. 1). The data recorded by the motion capture cameras include the vehicle's translational and rotational positions in the global frame $[X,Y,Z,\phi,\theta,\psi]^T$. Using the recorded orientation angles, the rotational matrix that links global frame to body frame $R_{ib}$ can be computed and thereby used in the dynamics model (1) in body frame. The velocities are calculated by differentiating the positions on each timestamp in the recorded data, and then passed through the rotational matrix to get those in body frame; note that angular velocities in body frame $[p,q,r]^T$ are achieved by applying the velocity rotational matrix

$$\begin{bmatrix}\dot\phi\\ \dot\theta\\ \dot\psi\end{bmatrix} = \begin{bmatrix}1 & s\phi t\theta & c\phi t\theta\\ 0 & c\phi & -s\phi\\ 0 & s\phi sc\theta & c\phi sc\theta\end{bmatrix}\begin{bmatrix}p\\ q\\ r\end{bmatrix} \quad (5)$$

using the notations $t\cdot$ as $tan(\cdot)$ and $sc\cdot$ as $sec(\cdot)$. The testing area in the diving pool was limited to be a $12 \times 14 \times 6\ m^3$ volume to capture the glider motion by the cameras. The global frame in the pool was defined using a calibration device of $Qualisys^{TM}$ as detailed in its user manual. The setup of the motion capture system and the glider in the diving pool is shown in Fig. 3.

## 2.3 Bayesian parameter estimation

The equation of motion of the glider system is essentially a nonlinear optimization problem, to find the best values of the parameters to minimize the difference between the recorded data and the estimated data. Thus, the optimization problem can be mathematically expressed as follows,

$$\tau_* = \mathrm{argmin} f(Y_{obeservation}, Y(\tau)_{simulation}) \quad (6) \quad \tau \in D$$

in which $\tau$ is the vector containing all the parameters to be estimated, $D$ is a domain of all the possible parameter values, and $f$ is an error function between the recorded experimental data ($Y^{observation}$) and the estimated data ($Y(\tau)^{simulation}$).

The equation of motions (EoM) (1) indicates that despite the complex form, the accelerations in the body frame are linearly responding to the hydrodynamics coefficients $K$'s for all axes. For such a distribution with a large number of unknown parameters, a method of calibration utilizes Bayesian sequential sampling methods known as the Markov Chain Monte Carlo (MCMC) method. In this work, we applied the Metropolis-Hastings algorithm (Hastings (1970); Metropolis et al. (2004)), which combines two major concepts to provide an easy but robust method:

- Transition function – The estimated parameter values need to be updated at each iteration in the algorithm, and the calculation of the new values is accomplished by the provision of a transition function. Because for each unit of the underwater glider, the hydrodynamics coefficients are expected to remain constant, regardless of the motions of the glider, it is reasonable to assume the parameter values to be in the form of normal distribution. Therefore, the updated values are calculated as $\tau_{new} = \tau_{current}+\xi$, $\xi \sim N(0,\sigma_{new})$, and the transition function, known as the random walk kernel, yields

$$Q(\tau_{new}|\tau_{current}) = \frac{1}{\sqrt{2\pi}\sigma_{new}}e^{-\frac{1}{2}(\frac{\tau_{new}-\tau_{current}}{\sigma_{new}})^2}$$

(7)

- Acceptance of proposed values – For each proposed set of values of estimated parameters, whether to keep them as the new values or repeat the sampling needs to be determined by calculating the acceptance probability given as

$$ap = \min(1, \frac{\Pi(\tau_{new})Q(\tau_{new}|\tau_{current})}{\Pi(\tau_{current})Q(\tau_{current}|\tau_{new})}) \quad (8)$$

in which $\Pi$ is the stationary distribution of the Markov chain. The proposed values are accepted as the new estimated parameter values for a probability of $ap$, and are rejected while the current parameter values are retained otherwise.

## 3. NONLINEAR CONTROLLER DESIGN

With the parameters estimated, the numeric dynamics model can be calculated by plugging the estimated values in the equation of motion (1). The glider system is a highly nonlinear system with potential modeling errors and uncertainties. Therefore, the methodology of deterministic robust control is used to design a MIMO controller for ROUGHIE. This section presents the system model analysis and designed approach for depth, roll, and pitch control.

### 3.1 Model analysis

The first step of designing a controller is to rewrite the system model equations in Input/Output (I/O) form for better system analysis. There are three inputs in ROUGHIE: the angle of servo ($\gamma$), the moving distance of the linear mass ($\Delta r_s$), and the buoyancy mass ($m_b$) which determines the moving distance of the plunger in the tank ($\Delta r_b = \Delta r_b(m_b)$). The three corresponding outputs include the rolling angle ($\phi$), the pitching angle ($\theta$), and the depth ($Z$). Because underwater glider is a slow-varying system (Mahmoudian (2009)), the rate of change of the inertia matrix $\dot{M}$ is assumed to be zero for simplification of model analysis and controller design.

For the pitch motions, the EoM yields

$$\dot{\theta} = c\phi \cdot q - s\phi \cdot r \quad (9)$$

the derivative of which gives a second-order differential equation

$$\ddot{\theta} = c\phi \cdot \dot{q} - s\phi \cdot \dot{r} - (q \cdot s\phi + r \cdot c\phi)\dot{\phi} \quad (10)$$

Because the velocity rotational matrix (5) indicates that $\dot{\phi} = p + s\phi t\theta \cdot q + c\phi t\theta \cdot r$ (11) the second order equation (10) can be rewritten as

$$\ddot{\theta} = c\phi \cdot \dot{q} - s\phi \cdot \dot{r} - (q \cdot s\phi + r \cdot c\phi)(p + s\phi t\theta \cdot q + c\phi t\theta \cdot r) \quad (12)$$

and substituting the non-derivative terms with $A$ yields

$$\ddot{\theta} = c\phi \cdot \dot{q} - s\phi \cdot \dot{r} + A \quad (13)$$

Using the expressions of $\dot{q}$ and $\dot{r}$ in EoM and rewriting the position of linear mass on x-axis $r_{s1}$ as $r_{s1} = r_{sx0} + \Delta r_s$ where $r_{sx0}$ is the initial position of linear mass on the rail give the I/O form of equation of pitch motions

$$\ddot{\theta} = A + c\phi \cdot \frac{1}{I_{yy}}[(I_{zz} - I_{xx})pr - g(s\theta(\Sigma m_i r_{i3}) + c\phi c\theta(m_b r_{b1} + m_r r_{r1} + m_s r_{sx0})) + T_3]$$
$$- m_s g c\theta(\frac{1}{I_{yy}}c^2\phi + \frac{1}{I_{zz}}s^2\phi) \cdot \Delta r_s \quad (14)$$

Substituting the coefficient of $\Delta r_s$ as $C$ and any other terms as $B$ gives a feedback-linearized form of the equation $\ddot{\theta} = B + C \cdot \Delta r_s$ (15)

For the depth motions, using the rotational matrix and EoM, it is easy to find the rate of depth to be

$$\dot{Z} = -u \cdot s\theta + v \cdot s\phi c\theta + w \cdot c\phi c\theta \quad (16)$$

Taking the derivative of this first-order equation and plugging in the velocity rotational matrix (5) yields

$$\ddot{Z} = (p + s\phi t\theta \cdot q + c\phi t\theta \cdot r)(v \cdot c\phi c\theta - w \cdot s\phi c\theta)$$
$$- (c\phi \cdot q - s\phi \cdot r)(u \cdot c\theta + v \cdot s\phi s\theta + w \cdot c\phi s\theta) \quad (17)$$
$$- s\theta \cdot \dot{u} + s\phi c\theta \cdot \dot{v} + c\phi c\theta \cdot \dot{w}$$

Substituting the non-differentiated terms with $D$ and representing each of $\dot{u}$, $\dot{v}$ and $\dot{w}$ using EoM gives the I/O form of depth motions

$$\ddot{Z} = D - s\theta \cdot \frac{1}{m_t}[m_t(rv - qw) + F_1]$$
$$+ s\phi c\theta \cdot \frac{1}{m_t}[m_t(pw - ru) + F_2]$$
$$+ c\phi c\theta \cdot \frac{1}{m_t}[m_t(qu - pv) + F_3]$$
$$+ \frac{g}{m_t} \cdot m_b \quad (18)$$

Rewriting the coefficient of $m_b$ as $G$ and all the other terms as $E$ gives the feedback-linearized form

$$\ddot{Z} = E + G \cdot m_b \quad (19)$$

The roll motions have a kinematic equation of

$$\dot{\phi} = p + s\phi t\theta \cdot q + c\phi t\theta \cdot r \quad (20)$$

which leads to the rolling acceleration as

$$\ddot{\phi} = \dot{p} + c\phi t\theta \cdot q\dot{\phi} + s\phi sc^2\theta \cdot q\dot{\theta} + s\phi t\theta \cdot \dot{q}$$
$$- s\phi t\theta \cdot r\dot{\phi} + c\phi sc^2\theta \cdot r\dot{\theta} + c\phi t\theta \cdot \dot{r} \quad (21)$$

Replacing the non-derivative parts with $H$ and expanding the rest of the equation using EoM give the I/O form

$$\ddot{\phi} = H + \frac{1}{I_{xx}}[(I_{yy} - I_{zz})qr + T_1]$$
$$+ \frac{s\phi t\theta}{I_{yy}}[(I_{zz} - I_{xx})pr - g \cdot c\phi c\theta \cdot (\Sigma m_i r_{i1}) + T_2]$$
$$+ \frac{c\phi t\theta}{I_{zz}}[(I_{xx} - I_{yy})pq + g \cdot s\phi c\theta \cdot (\Sigma m_i r_{i1}) + T_3]$$
$$+ \frac{g \cdot c\theta}{I_{xx}}(-m_b\Delta r_b - m_r R - m_s \Delta r_s)sin(\gamma - \phi)$$
$$+ \frac{g \cdot s\phi t\theta s\theta}{I_{yy}}(m_b\Delta r_b + m_r R + m_s \Delta r_s) \cdot c\gamma$$
$$+ \frac{g \cdot c\phi t\theta s\theta}{I_{zz}}(-m_b\Delta r_b - m_r R - m_s \Delta r_s) \cdot s\gamma \quad (22)$$

By substituting the terms not having the input $\gamma$ with $J$, and according to trigonometry function identities, the equation is simplified to be

$$\ddot{\phi} = J + (-m_b\Delta r_b - m_r R - m_s \Delta r_s)g$$
$$\cdot [(\frac{c\theta c\phi}{I_{xx}} + \frac{c\phi t\theta s\theta}{I_{zz}})s\gamma - (\frac{c\theta s\phi}{I_{xx}} + \frac{s\phi t\theta s\theta}{I_{yy}})c\gamma] \quad (23)$$

Euler angles feedback including $\phi$ and $\theta$ are measured using an onboard IMU installed near the tail of ROUGHIE inside its hull. Depth feedback $Z$ is measured by an embedded pressure sensor on the beacon installed underneath the hull. The positions of individual components inside the hull including linear mass ($r_s$) and tank plunger ($r_b$) are valued by the draw-wire sensors attached to them, and the buoyancy mass ($m_b$) is estimated based on the tank's draw-wire sensor reading which gives the height of the approximate cylinder of water inside the tank, the bottom surface area of the plunger, and water density estimated by the pressure and temperature measured by embedded sensors on the beacon. Therefore, some system uncertainties may be caused by sensor noise, the precision of the sensors, and disturbance from the surrounding environment, etc.

### 3.2 Controller design

Because both depth (19) and pitch (15) motions have an I/O equation form of $\ddot{x}_i = f(x_i, x_j, x_k, \ldots) + g(x_i, x_j, x_k, \ldots) \cdot u_i$ (24) where x represents the corresponding state variables and u is the input, the controller design for these two sub-systems follows the same procedure using the Lyapunov function.

Define a system error $e = x - x_d$, where $x_d$ is the target value of the state variable, such that $\dot{e} = \dot{x} - \dot{x}_d$ and $\ddot{e} = \ddot{x} - \ddot{x}_d$. Considering all the parameters as slow-variant variables, and that the sub-system is second-ordered, the sliding surface is defined as

$$s = k_1 e + \dot{e}$$
$$\dot{s} = k_1 \dot{e} + \ddot{e}$$
(25)

where $k_1$ is a positive constant. Then a Lyapunov function is defined as,

$$V = \frac{1}{2} s^2$$
$$\dot{V} = s\dot{s} = s(k_1 \dot{e} - \ddot{x}_d + \ddot{x}) = s(k_1 \dot{e} - \ddot{x}_d + f + g \cdot u)$$
(26)

To handle the nonlinearity of the system, a sliding mode controller is developed as

$$u_{smc} = -k_2 sign(s) sign(g) \quad (27)$$

where $k_2$ is a positive constant. Plugging the sliding mode controller into Lyapunov function yields

$$\dot{V} = s(k_1 \dot{e} - \ddot{x}_d + f - g \cdot k_2 sign(s) sign(g))$$
$$= s(k_1 \dot{e} - \ddot{x}_d + f) - s(g \cdot k_2 sign(s) sign(g)) \quad (28)$$
$$\leq |s||k_1 \dot{e} - \ddot{x}_d + f| - k_2 |s||g|$$

To ensure stability of the system, $\dot{V} \leq 0$, which gives the range of controller coefficient $k_2$

$$k_2 \geq \left| \frac{k_1 \dot{e} - \ddot{x}_d + f}{g} \right|$$
(29)

To lower the influence of the chattering problem of sliding mode control Utkin and Lee (2006), a saturation function is defined to replace the sign function

$$u_{smc} = -k_2 sat(\frac{s}{\varepsilon}) sat(\frac{g}{\varepsilon})$$
(30)

where $0 < \varepsilon \ll 1$ and

$$sat(\cdot) = \begin{cases} \cdot & \text{if } |\cdot| \leq 1 \\ sign(\cdot) & \text{otherwise} \end{cases}$$
(31)

To reduce the tracking errors, a combination of a feedforward and a PD controller using backstepping method is added

$$u_{bsc} = \frac{1}{g}(-k_1 \dot{e} + \ddot{x}_d - f - k_3 s)$$
(32)

in which $k_3$ is a positive constant. Then the updated Lyapunov function with input $u = u_{smc} + u_{bsc}$ yields

$$\dot{V} = s(-k_1 \dot{e} + \ddot{x}_d - f - k_3 s$$
$$- g \cdot k_2 sat(s) sat(g) + k_1 \dot{e} - \ddot{x}_d + f) \quad (33) =$$
$$-k_3 s^2 - k_2 |s||g| \leq 0$$

As a result, the depth and pitch controllers are expressed as

- Depth control:
$$m_b = \frac{1}{G}(-k_{m1} \dot{e} + \ddot{Z}_d - E - k_{m3} s) - k_{m2} sat(s) sat(G)$$
(34)

in which $k_{m1}, k_{m2}, k_{m3} > 0$, $s = k_{m1} e + \dot{e}$, $e = Z - Z_d$, $E = (p + s\phi t\theta \cdot q + c\phi t\theta \cdot r)(v \cdot c\phi c\theta - w \cdot s\phi c\theta) - (c\phi \cdot q - s\phi \cdot r)(u \cdot c\theta + v \cdot s\phi s\theta + w \cdot c\phi s\theta) - s\theta \cdot \frac{1}{m_t}[m_t(rv - qw) + F_1] + s\phi c\theta \cdot \frac{1}{m_t}[m_t(pw - ru) + F_2] + c\phi c\theta \cdot \frac{1}{m_t}[m_t(qu - pv) + F_3]$, and $G = \frac{g}{m_t}$.

- Pitch control:
$$\Delta r_s = \frac{1}{C}(-k_{s1} \dot{e} + \ddot{\theta}_d - B - k_{s3} s) - k_{s2} sat(s) sat(C)$$
(35)

in which $k_{s1}, k_{s2}, k_{s3} > 0$, $s = k_{s1} e + \dot{e}$, $e = \theta - \theta_d$, $B = -(q \cdot s\phi + r \cdot c\phi)(p + s\phi t\theta \cdot q + c\phi t\theta \cdot r) + c\phi \cdot \frac{1}{I_{yy}}[(I_{zz} - I_{xx})pr - g(s\theta(\Sigma m_i r_{i3}) + c\phi c\theta(m_b r_{b1} + m_r r_{r1} + m_s r_{sx0})) + T_3$
$\frac{1}{I_{zz}} s^2 \phi)$.  ], and $C = -m_s g \cdot c\theta(\frac{1}{I_{yy}} c^2 \phi +$

such that the total Lyapunov function

$$V_t = \frac{1}{2} \Sigma s_i^2$$
$$\dot{V}_t = -k_{m3} s_m^2 - k_{m2} |s_m||G| - k_{s3} s_s^2 - k_{s2} |s_s||C|$$
$$\leq 0$$
(36)

indicates the stability of the control system. The controller structure contains a feedforward branch, a linear feedback branch, and a nonlinear feedback branch.

For roll motion control, due to the lack of real-time servo angle feedback, a feedforward/PD controller, same as presented in Ziaeefard et al. (2018), is applied to avoid the need of servo angle readings

$$\gamma = \phi_d + k_p e + k_d \dot{e} \quad (37)$$

where error is defined as $e = \phi_d - \phi$, $\dot{e} = \dot{\phi}_d - \dot{\phi}$.

## 4. SIMULATION AND RESULTS

This section presents the results of parameter estimation and controller simulations. In addition, the performance of a hybrid control system to achieve more complex maneuvers is demonstrated.

### 4.1 Parameter estimation

We collected experimental data for 45 gliding runs with approximately 900 observations each with our vehicle running with control input combinations of the following values: buoyancy mass $m_b$ at 1, 0.6, 0.2 times its maximum command value, sliding mass position $r_b$ at 1, 0.5, 0.1 times maximum value, and servo angle $\gamma$ at 1, 0.8, 0.6, 0.4, 0.2 times maximum value. The glider started with a small velocity in $u$- direction to fasten the process of initiating the flight process. The real-time positions of the glider were captured by the underwater cameras using its software (UI shown in Fig. 3) and then the recorded data were processed in MATLAB to find the necessary data like velocities and accelerations used in the EoM (1).

Parameters in the dynamic model were identified using experimental data including the the glider's orientations and positions with various combinations of control inputs, as mentioned in section 2. The MCMC method returns a normal distribution of values for each one of the parameters, as shown in Fig. 4. The figures, though showing multiple peaks for $K_L$, $K_p$, $K_q$, and $K_{MY}$, mostly show a normal distribution shape for each parameter. To minimize the loss of picking possible incorrect parameter values, only the mean values of each parameter estimated (as marked in red in Fig. 4) were picked to conduct the system control simulations.

### 4.2 Controller simulation

The proposed controller was used to simulate the underwater glider response to a variety of target pitch and depth positions under certain disturbances and modeling errors. The simulations were conducted as such using MATLAB/Simulink software: on all of the 6 degrees of freedom, the influence by disturbances and errors was added as white noise, i.e. $N(0, \sigma^2)$, to the acceleration terms in (1), and to push the controllers to their limitations, such influence was set to have high frequency (10 Hz) to simulate an extreme environment where highspeed underwater vortex or waves may be significantly influential; for pitch control, the vehicle was commanded to reach the target pitch angles of ±10°, ±30°, and ±45° while the target descending depth values below the initial position include 0 $m$, 2 $m$, 4 $m$, and 5 $m$; these values are picked because they are reachable in the diving pool and thus can be used for future experimental validations. To examine the performance of the proposed NLC, PID controllers for pitch and depth control were used as baselines, and their coefficients were determined using PID tuner in MATLAB to minimize the tuning bias. To prevent the possible control output overshoots due to large target output changes, a second-order filter was added before the constant target output to form a reference trajectory, s.t.
$ref = \frac{\omega_n^2}{s^2+2\zeta\omega_n+\omega_n^2} \cdot u_c$, so instead of directly reaching the target positions, the glider is tracking a preset trajectory on each controlled motion.

Fig. 5 illustrates the comparisons between NLC and PID on pitch control performances. The NLC is able to lower the impact of disturbance and noise as expected. PID results in an average tracking error of approximately 13% for larger target final values like shown in Fig. 5(iv) and Fig. 5(vi), while NLC, for the same scenario, leads to less than 5% tracking error in the output (Fig. 5(ii) and Fig. 5(v)). The final value of controlled output, however, is not precisely guaranteed compared to applying a PID controller, as shown in Fig. 5(i) and Fig. 5(v). So, the PID controller can be used as a supplement for NLC to ensure reaching a precise final value and NLC can be applied to reduce the influence of disturbance to improve the performance.

In Fig. 6 the performance of NLC and PID for tracking the desired trajectories is displayed. A successful ascending flight is not guaranteed using PID controller (Fig. 6(ii) and Fig. 6(vi)) due to the asymmetric dynamics of UG system (1). The results demonstrate that the PIDcontrolled flights fail to perform in the extreme conditions, however NLC is able to overcome the influence of rapid disturbances. In other cases, the tracking error in controlled output by NLC is lower the noise level in output and enhance the ascending flights.

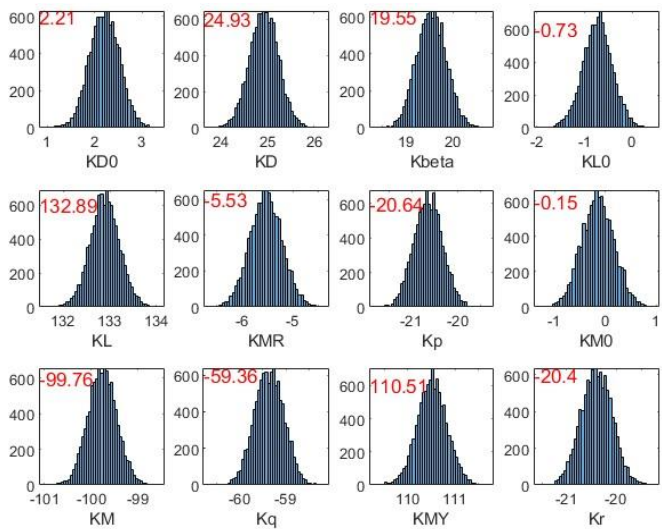

Fig.4.Parameter estimation results by MCMC method

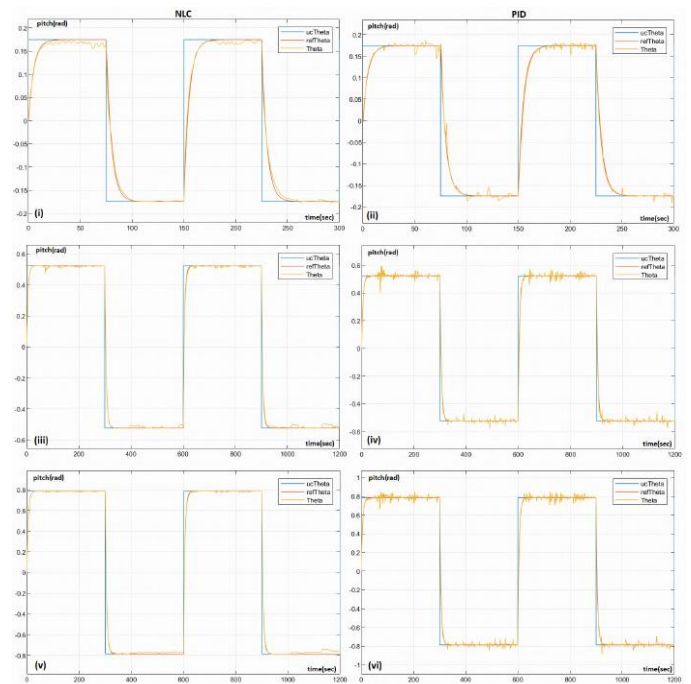

Fig. 5. NLC vs PID on pitch angle control

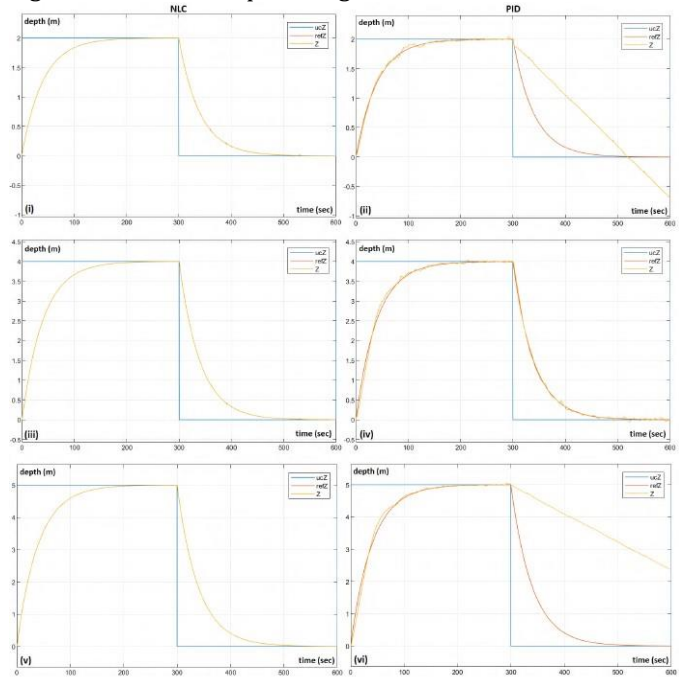

Fig. 6. NLC vs PID on depth control

Note that this work tested the controller in an extreme simulated environment in which rapid and random noises influence the motion of the vehicle. In reality, the water area where the glider is deployed is expected to have more regular and periodic but rapidly varying currents and waves as the disturbance. The proposed NLC is targeted to be applied in extreme water regions like, for example, fjord areas where the sea condition is usually relatively calm, but during the seasons of severe weather with strong winds, it becomes rough. Because for most of its flights, the glider is designed to be maintaining a steady flight rather than constantly changing gliding or heading angles, a threshold is added to the control system so that based on the impact of disturbance, the system can switch between NLC and PID to either use NLC to adjust the orientations, i.e. transitioning between steady flight segments, when the influence is too high, or otherwise use PID to simply maintain the steady flights.

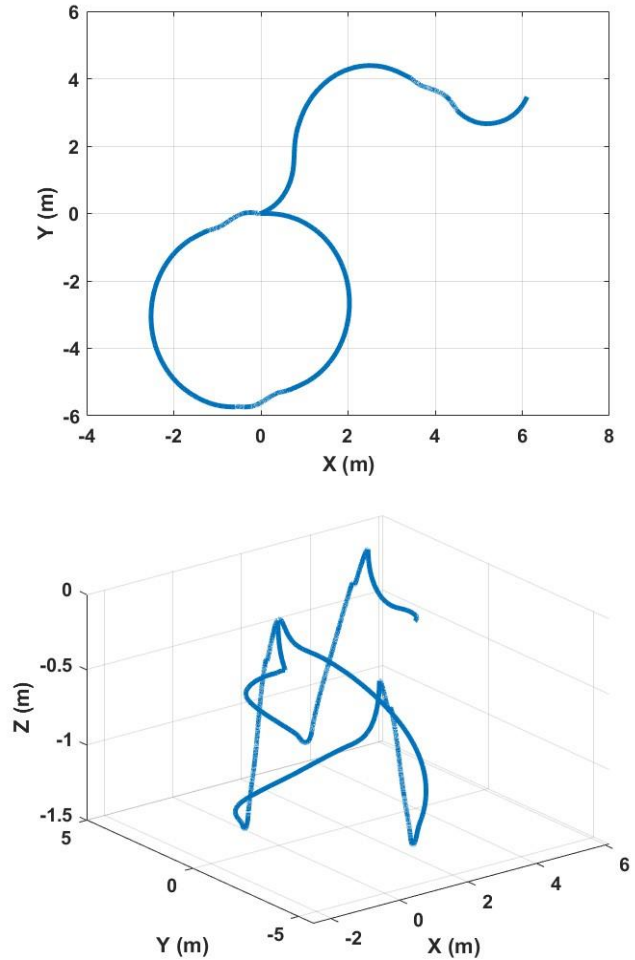

Fig. 7. A maneuvering transition between a complete circular pattern to an S-shaped pattern performed by the glider, illustrated in 2D top view and 3D view

Utilizing this controller switching system, a sample simulated *Circular to S-curved* maneuver is shown in Fig. 7, in which the glider, starting from the origin, transits to an S-shaped pattern after finishing a complete circle of gliding. The circular path is achieved by completing two cycles of descending+ascending glides while the roll angle is maintained at its maximum reachable value; the S-shaped path is accomplished by performing the same flight patterns but switching the direction of rolling during each cycle of gliding. Again, due to the extreme setup of the simulated disturbance and noise, some chattering is visible. This test shows that the proposed control system can connect steady flight segments with nonlinear transitions in between under environmental disturbance. The combination of the demonstrated flight patterns enables the glider to achieve more complex behaviors within constrained spaces and can be applied for underwater searching and navigating in regions with rough flow conditions.

## 5. CONCLUSION

In this work, the dynamics system of a custom-made underwater glider is identified using MCMC method involving experimental data collected by Qualisys underwater motion capture system. Then a nonlinear system controller is designed based on Lyapunov functions to overcome potential modeling errors and the expected disturbance in the environment in which the glider is deployed. The nonlinear controller is combined with PID controller to form a hybrid control system for the glider to efficiently perform more complex underwater maneuvers. The model and controller may serve as starting points for higher-level controls and path planning that account for complicated flow fields and underwater terrains.